\documentclass{article}
\usepackage{spconf,amsmath,graphicx,hyperref}
\usepackage{booktabs}

\title{PREPARED MIND, FAST RESPONSE: A TEMPORAL DECOUPLING FRAMEWORK FOR ADAPTIVE KNOWLEDGE ORCHESTRATION IN OPEN-DOMAIN DIALOGUE}

\name{Jinling Gan, Churong Liang, Runnan Li$^{\ast}$ \thanks{*Corresponding author}}
\address{
  Beijing University of Posts and Telecommunications, Beijing, China\\
  randown@bupt.edu.cn, liangchurong@bupt.edu.cn, runnan.li@bupt.edu.cn
  }
  
\begin{document}

%
\maketitle
\begin{abstract}

The latency-quality tradeoff is a fundamental constraint in open-domain dialogue AI systems, since comprehensive knowledge access necessitates prohibitive response delays. Contemporary approaches offer two inadequate solutions: lightweight instruct models achieve sub-second latency but lack reasoning depth, while tool-augmented ReAct agents enhance factuality through external knowledge at the cost of synchronous execution that blocks interaction during retrieval processes. \textbf{PMFR} is thus proposed, with a temporal decoupling framework that fundamentally resolves the contradiction through asynchronous knowledge orchestration. PMFR employs three coordinated components: (1) a Knowledge Adequacy Evaluator for real-time sufficiency assessment, (2) a Lightweight Response Generator for immediate user interaction, and (3) an Asynchronous Knowledge Refinement Agent for background knowledge enhancement. This architecture maintains continuous conversational flow while progressively enriching knowledge coverage through intelligent triggering mechanisms. Evaluation results on TopiOCQA demonstrate PMFR outperforms brute-force scaling: PMFR achieves \textbf{95.3\% latency reduction} (23.38s→1.09s) while preserving response quality comparable to heavyweight synchronous baselines (GEval-C: \textbf{0.613} vs. \textbf{0.620}).


\end{abstract}

\section{Introduction}

Building open-domain conversational AI represents a fundamental challenge in artificial intelligence, requiring systems to engage in natural, knowledge-grounded dialogue across unlimited topics while maintaining real-time responsiveness. As conversational AI becomes increasingly integrated into critical applications, from virtual assistants to customer service and educational platforms, the demand for systems that can seamlessly balance comprehensive knowledge access with interactive responsiveness has intensified dramatically.

Two distinct paradigms have established in large language models for dialogue systems. First, lightweight \emph{instruct} models\cite{ouyang2022instructgpt} respond in milliseconds but suffer from shallow reasoning and limited knowledge coverage. Second, \emph{thinking} variants incorporating explicit chain-of-thought reasoning~\cite{wei2022chainofthought,kojima2022zeroshot} significantly enhance deliberation quality at the cost of prohibitive per-turn latency, often exceeding 20 seconds per response. Recent technical reports formalize this dichotomy by releasing both model types and introducing \emph{thinking budgets} to manage computational trade-offs~\cite{yang2025qwen3technicalreport}, and Talker-Reasoner frameworks and DPT-Agent~\cite{christakopoulou2024agents,zhang-etal-2025-leveraging} attempt to address these limitations with dual-process architectures. However, in knowledge-intensive multi-turn dialogue scenarios, these approaches still block user interaction until retrieval completes, failing to resolve the core latency-quality contradiction.

\textbf{PMFR (Prepared Mind, Fast Response)}, a temporal decoupling framework to fundamentally resolve the latency-quality tradeoff through asynchronous knowledge orchestration, is thus proposed. PMFR separates immediate response generation from knowledge acquisition: a lightweight dialogue model coupled with a dynamic knowledge base produces sub-second responses, while an intelligent \emph{knowledge sufficiency monitor} determines when to asynchronously trigger a heavyweight ReAct-style agent for background knowledge enhancement. PMFR maintains continuous user interaction while progressively enriching knowledge coverage, effectively transforming traditional blocking operations into productive background learning. The primary contributions of the proposed work can be summarized as:

\begin{itemize}
\item \textbf{Temporal decoupling architecture}: PMFR is a framework that completely decouples user-facing response generation from knowledge acquisition, achieving 95.3\% latency reduction while preserving response quality through heterogeneous model collaboration.
\item \textbf{Adaptive knowledge orchestration}: An intelligent sufficiency-based gating mechanism that triggers external retrieval when necessary, optimizing computational efficiency through real-time knowledge assessment.
\item \textbf{Empirical validation}: Evaluation experiments have stated the effectiveness of the proposed PMFR, achieving comparable quality to large-scale CoT-augmented ReAct agents (0.613 vs. 0.620) with 21× faster response times (1.09s vs. 23.38s) on TopiOCQA bench.
\end{itemize}

\section{Related Work}
\subsection{Latency-Quality Tradeoffs in Dialogue Systems}

The latency-quality contradiction in dialogue systems represents a well-established challenge that has driven significant research efforts. Early systems like BlenderBot established the foundation by prioritizing interactive responsiveness~\cite{shuster2022blenderbot3}, while evaluation benchmarks including TopiOCQA, QReCC, and TREC CAsT formalized the requirements for multi-turn, knowledge-intensive interactions~\cite{adlakha2022topiocqa,anantha-etal-2021-open,choi2018quac,dalton2021cast,owoicho2022cast}. Contemporary large language model families explicitly acknowledge this tradeoff through dual-variant architectures: (i) lightweight \emph{instruct} models\cite{ouyang2022instructgpt} are able to achieve sub-second response times but exhibit limited reasoning depth and knowledge coverage; (ii) \emph{thinking} variants that incorporate explicit chain-of-thought reasoning~\cite{wei2022chainofthought,kojima2022zeroshot} to enhance deliberation quality at the cost of prohibitive latency. Recent technical reports formalize this dichotomy by introducing \emph{thinking budgets} as a mechanism for managing computational trade-offs under latency constraints~\cite{yang2025qwen3technicalreport}, highlighting the persistent challenge of balancing interactive responsiveness with reasoning sophistication.

\subsection{Temporal Dual-Process Architectures}

Dual-process dialogue systems draw inspiration from cognitive theories distinguishing fast, intuitive responses (System-1) from slow, deliberative reasoning (System-2). The Talker-Reasoner framework~\cite{christakopoulou2024agents} formalizes this separation through a lightweight talker component for immediate interaction and a heavyweight reasoner for complex problem-solving. Similarly, DPT-Agent~\cite{zhang-etal-2025-leveraging} implements System-1/System-2 coordination through finite-state control and theory-of-mind modules operating at different temporal scales.  And engineering approaches such as LiveMind, StreamingLLM, speculative decoding, and CALM reduce perceived or computational latency\cite{chen2024livemind,xiao2024streamingllm,schuster2022calm}; multi-agent and reflective methods such as AutoGen, CAMEL, AgentVerse, and Reflexion enhance robustness across tasks~\cite{wu2023autogen,li2023camel,chen2023agentverse,shinn2023reflexion}. 
However, these dual-process systems primarily address internal reasoning separation without resolving the fundamental knowledge acquisition challenge of external knowledge acquisition, and lack sophisticated mechanisms for knowledge sufficiency assessment.

\subsection{Adaptive Knowledge Orchestration}

Multi-turn open-domain dialogue presents complex knowledge orchestration challenges that extend beyond single-query optimization. Evaluation benchmarks, particularly TopiOCQA~\cite{anantha-etal-2021-open}, demonstrate that 28\% of conversational turns require accessing different knowledge sources as topics shift, while conversational retrieval systems like ConvDR and ORConvQA highlight the difficulty of maintaining coherence across topic transitions~\cite{yu2021convdr,mo2024haconvdr,qu2020orconvqa,li2023gcoqa,vu2023freshllms}.

\begin{figure*}[t]
  \centering
  \includegraphics[width=\linewidth]{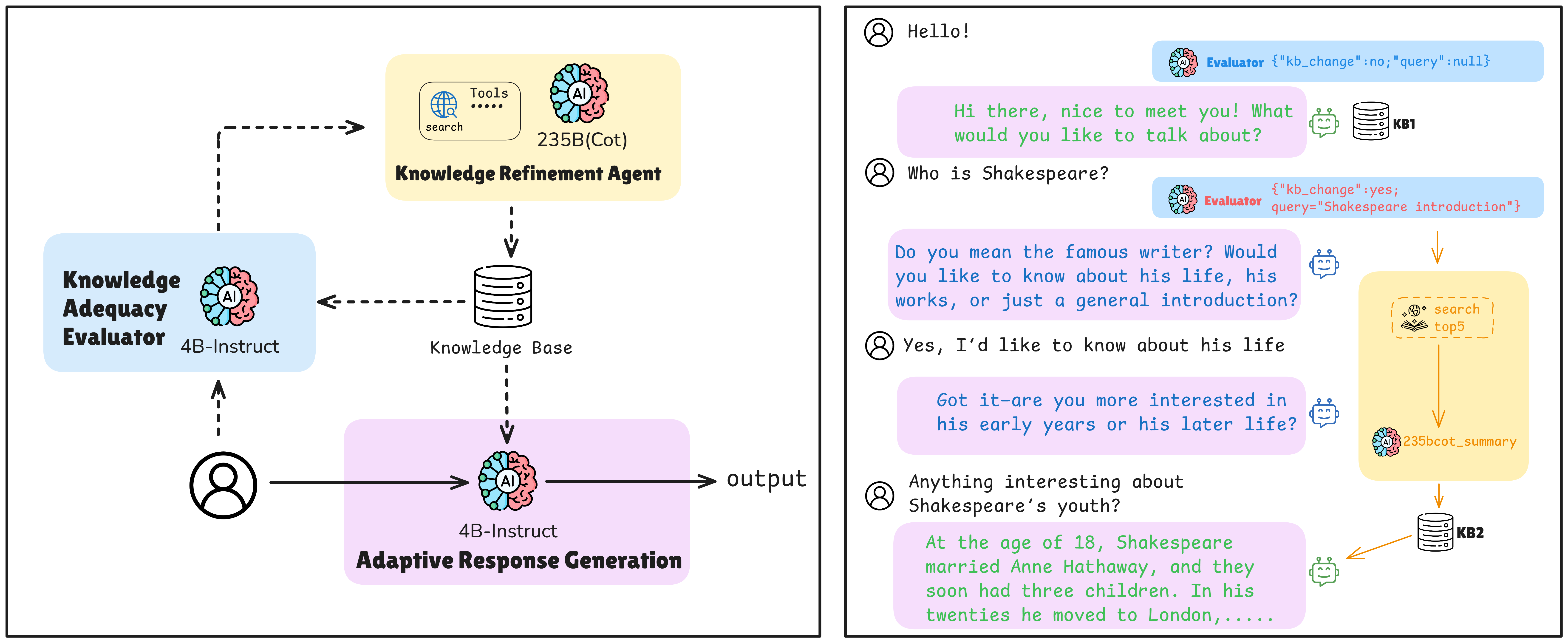}
  \caption{Architecture of the proposed framework is shown on the left side. On the right is a real case study demonstrating how the system evaluates knowledge sufficiency, triggers asynchronous knowledge refinement, and maintains dialogue flow.}
  
  \label{fig:model}
  \vspace{-3mm}
\end{figure*}

\section{Methodology}

\subsection{Problem Formulation}
We formalize open-domain dialogue as a sequential decision-making task where at each turn $t$, the system generates a response $r_t$ given user query $q_t$, dialogue history $H_{t-1} = \{(q_i, r_i)\}_{i=1}^{t-1}$, and knowledge base $K_t$. The fundamental challenge lies in optimizing the multi-objective function:

\begin{equation}
\mathcal{L}(r_t) = -\alpha \cdot Q(r_t | q_t, H_{t-1}) + \beta \cdot L(r_t) + \gamma \cdot C(r_t)
\end{equation}

\noindent where $Q(\cdot)$ measures response quality, $L(\cdot)$ represents response latency, $C(\cdot)$ denotes computational cost, and $\alpha, \beta, \gamma > 0$ are weighting parameters. Traditional approaches struggle with this optimization as improving quality typically requires increased computation and latency.

PMFR reformulates this by separating immediate response generation from knowledge enhancement:
\begin{align}
r_t &= f_{\text{fast}}(q_t, H_{t-1}, K_t) \label{eq:fast_response}\\
K_{t+1} &= \text{async}\{f_{\text{slow}}(q_t, H_{t-1}, K_t)\} \label{eq:async_update}
\end{align}

\noindent This decomposition enables sub-second response generation while maintaining knowledge coverage in background.

\subsection{PREPARED MIND, FAST RESPONSE}

PMFR, as shown in Fig.\ref{fig:model}, orchestrates three specialized components operating on distinct temporal scales: (i) a Knowledge Adequacy Evaluator $\mathcal{E}$ for real-time sufficiency assessment, (ii) a Lightweight Response Generator $\mathcal{G}$ for immediate user interaction, and (iii) an Asynchronous Knowledge Refinement Agent $\mathcal{A}$ for background knowledge enhancement. This architecture maintains interactive responsiveness while progressively enriching knowledge coverage.

The \textbf{Knowledge Adequacy Evaluator} $\mathcal{E}$ functions as a lightweight decision module that jointly performs sufficiency assessment and query contextualization. Given the current query $q_t$, dialogue history $H_{t-1}$, and knowledge base $K_t$, it produces a binary adequacy signal:

\begin{equation}
s_t = \mathcal{E}(q_t, H_{t-1}, K_t), \quad s_t \in \{0,1\}
\end{equation}

\noindent where $s_t=1$ indicates that the knowledge base $K_t$ is insufficient and requires refinement, while $s_t=0$ denotes adequacy. Asynchronous knowledge enhancement is triggered only when $s_t=1$, ensuring that updates occur selectively to balance efficiency with reliability.

The evaluator concurrently outputs a context-resolved reformulation $\tilde{q}_t$ to clarify entities and temporal references.

\begin{equation}
\tilde{q_t} = \text{decode}(W_c \cdot [\text{emb}(q_t), \text{emb}(\hat{H}{t-1})] + b_c)
\end{equation}

\noindent This reformulation improves retrieval accuracy and reduces update latency by aligning subsequent knowledge refinement with the precise informational needs of the dialogue context.

\subsection{Adaptive Response with Async Refinement}

Response generation in PMFR is guided by the evaluator’s adequacy decision. The system adapts through two modes:

\textbf{Direct Mode (KB-Hit).}
When the KB is adequate, the lightweight model responds directly with sub-second latency, ensuring responsiveness in well-covered domains.

\textbf{KB-Miss Transition Mode.}
When the KB is incomplete, the lightweight model sustains a brief, user-friendly dialogue to preserve naturalness and confirm query intent, while in parallel a heavyweight agent asynchronously acquires and refines external knowledge. This effectively turns “waiting” into “learning”: the system remains engaged, actively gathers contextual cues, and sharpens intent understanding, so that once knowledge is integrated, responses are richer, more reliable, and better aligned with user intent.

\textbf{Asynchronous Refinement Pipeline.} 
The heavyweight agent operates in background threads through three stages:
\begin{itemize}
  \item \textbf{Knowledge Acquisition.} Reformulated queries drive external search over web APIs, document repositories, and structured knowledge bases.
  \item \textbf{Evidence Reasoning.} Retrieved information is consolidated via chain-of-thought reasoning to extract salient facts and resolve contradictions.
  \item \textbf{Synopsis and Caching.} Distilled, provenance-aware summaries are cached for rapid reuse with attribution and confidence scores.
\end{itemize}

This design unifies immediate dialogue with asynchronous refinement, scaling computation by knowledge gaps rather than dialogue volume, and ensuring both responsiveness and progressive enrichment.

\begin{table*}[t]
\centering
\caption{Comprehensive evaluation of quality, latency, and stability. 
GEval-C evaluates turn-level correctness, GEval-RC assesses dialogue consistency and role preservation, 
Latency reports the mean response time, and P95 Latency represents the 95th-percentile worst-case performance relevant for deployment.}

\label{tab:main_results}

\begin{tabular}{lcccc}
\hline

\toprule
\textbf{Method}              & \multicolumn{1}{c}{\textbf{GEval-C}} & \multicolumn{1}{c}{\textbf{GEval-RC}} & \multicolumn{1}{c}{\textbf{Latency(s)}} & \multicolumn{1}{c}{\textbf{P95 Latency.}} \\  \hline
\textbf{Qwen-4B (Instruct, no tools)}             & 0.481                                & 0.595                                 & 1.155                                   & 1.844                                          \\
\textbf{Qwen-4B (CoT, no tools)}         & 0.511                                & 0.653                                 & 8.710                                   & 20.137                                         \\
\textbf{ReAct Agent (Backbone=Qwen-4B with CoT)}   & 0.460                                & 0.437                                 & 13.668                                  & 28.515                                         \\
\textbf{ReAct Agent (Backbone=Qwen-30B with CoT)}  & 0.573                                & 0.585                                 & 20.981                                  & 32.012                                         \\
\textbf{ReAct Agent (Backbone=Qwen-235B with CoT)} & 0.620                      & \textbf{0.845}                        & 23.375                                  & 49.443                                         \\
\textbf{PMFR (Ours)}                & \textbf{0.613}                       & 0.645                        & \textbf{1.090}                          & \textbf{1.810}                                 \\
 \hline

\end{tabular}
\end{table*}

\section{Experiments}

\textbf{Experimental Objectives.} We address three questions: (RQ1) Can temporal decoupling preserve response quality while reducing latency? (RQ2) Does architectural innovation outperform brute-force scaling in resolving the latency–quality tradeoff? (RQ3) What factors determine stability and deployment viability?

\subsection{Experimental Setup}
\textbf{Dataset.} We evaluate PMFR on \emph{TopiOCQA}~\cite{adlakha2022topiocqa}, which contains 2,514 turns across 205 sessions. Its hallmark is cross-topic follow-ups that demand both contextual reasoning and external knowledge, making it a rigorous testbed for latency–quality evaluation.  

\textbf{Protocol.} Since some queries can be answered from history while others require retrieval, TopiOCQA naturally tests dynamic knowledge integration. We adopt a strict zero-shot protocol with no task-specific fine-tuning, ensuring evaluation of architectural generalization under realistic deployment.


\textbf{Configuration.} Qwen3-4B\cite{yang2025qwen3technicalreport} runs locally on an RTX 4090 GPU (edge), while Qwen3-30B and Qwen3-235B are accessed via cloud APIs~\cite{qwen3}. All models adopt deterministic decoding ($T{=}0$), and latency reflects both computation and network delays.

\textbf{Evaluation.} 
\textbf{Quality.} We use DeepEval~\cite{liu2023gevalnlgevaluationusing}'s LLM-as-Judge framework for consistent assessment across all variants. To ensure reproducibility, both judge and evaluated systems use deterministic decoding ($T{=}0$). Two metrics are reported: (i) \emph{GEval-C}, measuring turn-level correctness against TopiOCQA (answer relevance $>$ factual accuracy $>$ style), addressing RQ1; and (ii) \emph{GEval-RC}, evaluating dialogue-level consistency (role adherence, history retention, contradiction avoidance), addressing RQ3.  

\textbf{Latency.} Under deployment-realistic constraints, we use non-streaming evaluation to reflect TTS scenarios where complete responses must precede synthesis. Metrics include: (i) mean response time for typical performance; (ii) 95th percentile (P95) latency for worst-case analysis; and (iii) a 10-second inter-turn interval to simulate realistic user thinking time and avoid artificial inflation from rapid queries.

\subsection{Baseline Systems}

We compare PMFR against three paradigms:  

\textbf{Direct Response.} \emph{Qwen-4B Instruct} provides immediate answers with minimal latency but limited reasoning capability, while \emph{Qwen-4B CoT} augments with chain-of-thought reasoning, improving quality at higher latency.  

\textbf{Synchronous Tool-Augmented.} \emph{ReAct Agents} (4B, 30B, 235B) interleave reasoning and retrieval, enabling systematic analysis of scaling effects on the latency–quality tradeoff.  

\textbf{Proposed.} \emph{PMFR} decouples response and knowledge enhancement: a 4B generator handles immediate interaction using $K_t$, while a 235B asynchronous agent enriches knowledge in the background. This architecture enables direct comparison with both lightweight (4B) and heavyweight (235B) baselines while maintaining the computational resources of the strongest synchronous approach.

\subsection{Experimental Results}

\textbf{Main Findings.}  
Table~\ref{tab:main_results}reports three main conclusions:
(1) temporal decoupling achieves near-optimal quality with sharply reduced latency;  
(2) architectural design yields more benefit than brute-force scaling;  
(3) stability is more critical than peak accuracy for practical deployment.

\textbf{Topline (RQ1–RQ3).}
(1) \emph{Quality vs.\ Latency.} PMFR retains the correctness of the strongest synchronous system (GEval-C $0.613$ vs.\ $0.620$) while reducing mean latency from $23.38$\,s to $1.09$\,s ($\approx21\times$, $-95\%$), setting a new Pareto point where factuality and responsiveness coexist. 
(2) \emph{Design vs.\ Scale.} Scaling ReAct from 4B$\rightarrow$235B improves GEval-C by $+34.8\%$ but also increases latency from $13.67$\,s to $23.38$\,s; in contrast, PMFR matches large-scale quality with consistent sub-2\,s latency, indicating architectural decoupling outperforms brute-force scaling. 
(3) \emph{Stability.} PMFR’s latency distribution remains tight (P95 $1.81$\,s, $+66\%$ over mean), whereas 235B ReAct exhibits heavy tails (mean $23.38$\,s, P95 $49.44$\,s, $+111\%$), which undermines real-time UX.

\textbf{Ablation and Component Analysis.}
The baseline progression highlights two extremes. At the low end, \emph{Qwen-4B} delivers the fastest responses (1.15\,s) but limited reasoning quality. At the high end, \emph{ReAct with a 235B backbone} achieves the strongest correctness (GEval-C $0.620$, GEval-RC $0.845$) but suffers prohibitive latency (23.38\,s mean, 49.44\,s P95). Scaling between them (4B$\rightarrow$30B$\rightarrow$235B) improves quality but latency grows superlinearly, confirming that brute-force scaling cannot resolve the trade-off. 
\emph{PMFR} bridges these regimes by pairing a fast 4B generator with a strong 235B slowline, delivering near-optimal quality with sub-2\,s latency.  

 \textbf{Analytical Insights.}
From our evaluation, four principles emerge:  
(1) Explicit reasoning raises quality but also induces prohibitive delays, making temporal decoupling essential.  
(2) Tool orchestration only works with sufficiently large backbones; small models collapse under retrieval overhead.  
(3) Brute-force scaling quickly saturates—beyond 30B, quality gains plateau while latency grows superlinearly.  
(4) Stability is as critical as mean latency, and only PMFR sustains reliable sub-2s responses.


\section{Conclusion}
This work presents \textbf{PMFR}, a temporal decoupling framework that resolves the latency--quality trade-off in conversational AI. By separating lightweight responses from asynchronous knowledge refinement, PMFR preserves \textbf{98.9\%} correctness of heavyweight baselines (0.613 vs. 0.620 GEval-C) while reducing latency by \textbf{95.3\%} (1.09s vs. 23.38s) and maintaining stable sub-2s responses (P95: 1.81s). Beyond empirical gains, PMFR establishes a generalizable paradigm---adequacy-driven retrieval, background refinement, and temporal separation of concerns---for balancing responsiveness and knowledge in real-time intelligent systems.

\bibliographystyle{IEEEbib}
\bibliography{icassp}

\end{document}